\DocumentMetadata{}
\documentclass[acmtog]{acmart}
\acmSubmissionID{169}
\usepackage{booktabs} 

\usepackage{mathtools}
\usepackage{bm}
\usepackage{float}
\usepackage{graphicx}
\usepackage{color, colortbl}
\usepackage{multirow}
\usepackage{subcaption}
\usepackage{multirow}
\usepackage{bbm}
\usepackage[ruled]{algorithm2e} 

\usepackage[]{xcolor}
\SetAlFnt{\small}
\SetAlCapFnt{\small}
\SetAlCapNameFnt{\small}
\SetAlCapHSkip{0pt}

\copyrightyear{2024}
\acmYear{2024}
\setcopyright{acmlicensed}\acmConference[SA Conference Papers '24]{SIGGRAPH Asia 2024 Conference Papers}{December 3--6, 2024}{Tokyo, Japan}
\acmBooktitle{SIGGRAPH Asia 2024 Conference Papers (SA Conference Papers '24), December 3--6, 2024, Tokyo, Japan}
\acmDOI{10.1145/3680528.3687563}
\acmISBN{979-8-4007-1131-2/24/12}

\makeatletter
\DeclareRobustCommand\onedot{\futurelet\@let@token\@onedot}
\def\@onedot{\ifx\@let@token.\else.\null\fi\xspace}

\newcommand{\figref}[1]{Fig.~\ref{#1}}
\newcommand{\secref}[1]{Section~\ref{#1}}

\newcommand{\tabref}[1]{Tab.~\ref{#1}}
\newcommand{\update}[1]{{\color{black}{#1}}}
\newcommand{\updatered}[1]{{\color{black}{#1}}}
\newcommand{\updategreen}[1]{{\color{black}{#1}}}

\newif\ifshowcomments

\ifshowcomments
        \newcommand{\note}[3]{{\textcolor{#2}{[\textbf{#1: #3}]}}}
        
        \newcommand{\Shreyas}[1]{\note{Shreyas}{cyan}{#1}} 
	\newcommand{\Sammy}[1]{\note{Sammy}{teal}{#1}}
        
        \newcommand{\Edo}[1]{\note{Edo}{blue}{#1}}
        
	\newcommand{\Fadime}[1]{\note{Fadime}{magenta}{#1}}
        
        \newcommand{\Bugra}[1]{\note{Bugra}{orange}{#1}}
        
        \newcommand{\Tomas}[1]{\note{Tomas}{red}{#1}}
        
\else
        \newcommand{\note}[3]{\unskip}
        \newcommand{\Shreyas}[1]{\unskip}
	\newcommand{\Edo}[1]{\unskip}
	\newcommand{\Fadime}[1]{\unskip}
	\newcommand{\Bugra}[1]{\unskip}
        \newcommand{\Tomas}[1]{\unskip}
	\newcommand{\Sammy}[1]{\unskip}
\fi

\newcommand{\methodname}{DiffH$_{2}$O}

\newcommand{\TITLE}{\methodname: Diffusion-Based Synthesis of Hand-Object Interactions from Textual Descriptions}

\newcommand{\colorRef}[1]{\textcolor{black}{#1}} 

\newcommand{\refsec}[1]{\colorRef{Sec.~\ref{#1}}}

\newcommand{\heading}[1]{\textbf{#1}.}

\definecolor{Gray}{gray}{0.9}

\newcommand{\R}{\rm I\!R}

\renewcommand{\thefootnote}{\fnsymbol{footnote}}

\makeatletter
\def\and{
  \end{tabular}%
  \hskip 2.85em \@plus.17fil%
  \begin{tabular}[t]{c}}
\makeatother

\begin{document}
\title{\TITLE}

\author{Sammy Christen}
\email{sammychristen@gmail.com}
\affiliation{\institution{ETH}\country{Switzerland}}
\affiliation{\institution{Meta}\country{Switzerland}}
\authornote{This work was done during an internship at Meta.}

\author{Shreyas Hampali}
\email{hampali@meta.com}
\affiliation{\institution{Meta}\country{Switzerland}}

\author{Fadime Sener}
\email{famesener@meta.com}
\affiliation{\institution{Meta}\country{Switzerland}}

\author{Edoardo Remelli}
\email{edoremelli@meta.com}
\affiliation{\institution{Meta}\country{Switzerland}}

\author{Tomas Hodan}
\email{tomhodan@meta.com}
\affiliation{\institution{Meta}\country{Switzerland}}

\author{Eric Sauser}
\email{esauser@meta.com}
\affiliation{\institution{Meta}\country{Switzerland}}

\author{Shugao Ma}
\email{shugao@meta.com}
\affiliation{\institution{Meta}\country{United States of America}}

\author{Bugra Tekin}
\email{bugratekin@meta.com}
\affiliation{\institution{Meta}\country{Switzerland}}


%
%

\renewcommand{\shortauthors}{S. Christen, S. Hampali, F. Sener, E. Remelli, T. Hodan, E. Sauser, S. Ma, B. Tekin}

\begin{abstract}

We introduce \methodname, a new diffusion-based framework for synthesizing realistic, dexterous hand-object interactions from natural language. Our model employs a temporal two-stage diffusion process, dividing hand-object motion generation into grasping and interaction stages to enhance generalization to various object shapes and textual prompts. To improve generalization to unseen objects and increase output controllability, we propose grasp guidance, which directs the diffusion model towards a target grasp, seamlessly connecting the grasping and interaction stages through a motion imputation mechanism.  We demonstrate the practical value of grasp guidance using hand poses extracted from images or grasp synthesis methods. Additionally, we provide detailed textual descriptions for the GRAB dataset, enabling fine-grained text-based control of the model output. Our quantitative and qualitative evaluations show that~\methodname~generates realistic hand-object motions from natural language, generalizes to unseen objects, and significantly outperforms existing methods on a standard benchmark and in perceptual studies.

\end{abstract}



\begin{teaserfigure}
\centering
\includegraphics[width=0.86\textwidth]
{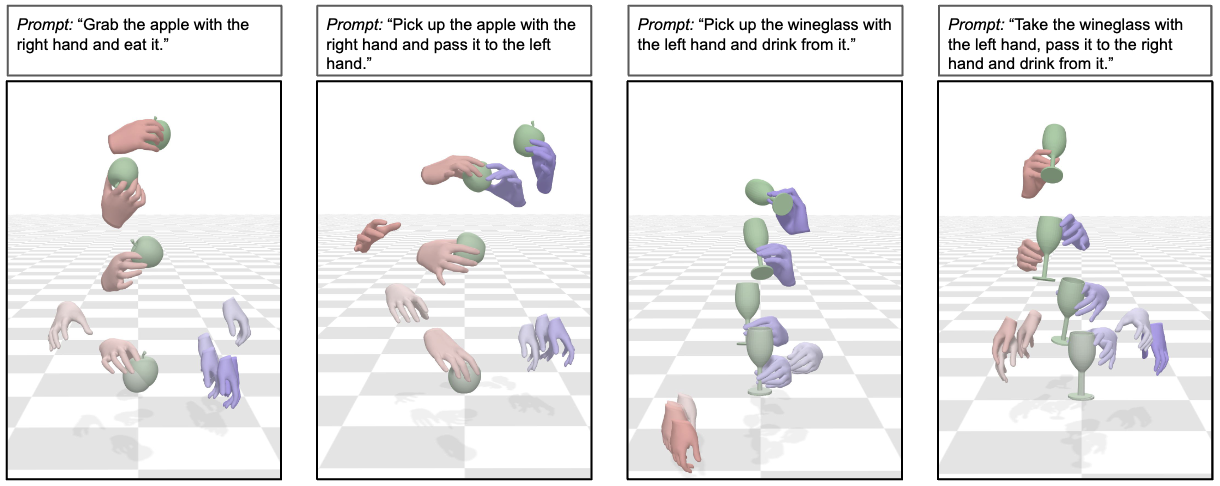}
\caption{We introduce \methodname, a diffusion-based framework to synthesize dexterous hand-object  interactions. \methodname~generates realistic hand-object motion from natural language, generalizes to unseen objects at test time and enables fine-grained control over the motion with detailed textual descriptions. Time is visualized with a color code where lighter shades denote the past. Best seen in the supplemental video.}
\label{fig:teaser}
\end{teaserfigure}

\maketitle

\section{Introduction}
\label{introduction}

Modeling and understanding hand-object interactions (HOI) is an important and crucial task to empower machines to interact with and assist humans. Being able to seamlessly generate hand-object motions holds the promise to enable synthetic data generation \cite{leonardi2023synthetic}, assist robots in training human-robot interactions in simulation, or enhance realism in virtual manipulation experiences. For example, in virtual reality (VR), hand interactions still often rely on heuristics and controllers that simply attach objects to the hand according to pre-defined grasps. Being able to faithfully produce object manipulation based on an input signal such as text or a few past frames of hand and object poses could largely increase the immersiveness of such interactions. 

Generating realistic hand-object interactions in 3D is challenging as the resulting motions are required to be plausible in several different aspects. First, the motions must be plausible in terms of \emph{geometry}, such that hand and object intersections are minimized and the grasp appears stable. Second, the motions must be plausible in terms of \emph{semantics}, such as the hands respect natural object affordances (e.g., a cup is grasped by its handle and not flipped upside down). Third, the motions must be plausible in terms of \emph{time}, such as the hand and object motions are synchronized and the dynamics appear natural.
Another challenge in generating HOI comes from the limited scale of existing hand-object datasets, which are around 10x smaller than human motion datasets~\cite{AMASS} and 1000x smaller than image datasets~\cite{deng2009imagenet}.

Several recent works have successfully leveraged diffusion models for full-body human motion generation
but either neglect objects ~\cite{tevet2023mdm, karunratanakul2023gmd} or focus only on coarse motion of larger objects (e.g., a chair) without finger predictions \cite{xu2023interdiff}. Another method, IMoS~\cite{IMoS}, generates HOI with smaller objects (e.g., a cup or stapler) by first predicting upper-body motions and then optimizing object poses in post-processing. However, IMoS assumes that the object is in-hand from the beginning, handles only objects seen during training, and, as shown in our experiments, often yields artifacts (e.g., hand-object interpenetration or non-plausible contacs).
In this paper, we aim to learn a model that \emph{generates natural, fine-grained hand-object interactions and generalize to objects unseen during training}.

Towards this goal, we contribute a novel diffusion-based framework, \methodname, that generates hand-object interactions based on text prompts and geometry of the object. We propose several techniques to deal with the challenges of data scarcity and object generalization.
\textbf{Decoupling Grasping and Interaction:} Hand-object interactions can be invariably split into a grasping phase where the hand(s) approaches the static object for grasping, and an interaction phase where the object is manipulated based on an action intent. For example, drinking from a cup can be performed with multiple grasps, and given a grasp, multiple actions are possible. We use this simple observation to obtain a grasping and an interaction diffusion model, and introduce an inpainting technique called subsequence imputing to allow continuity between the two outputs. In contrast to previous diffusion frameworks that employ coarse-to-fine (trajectory-to-full motion) motion generation \cite{karunratanakul2023gmd}, we split the task temporally and propose a  temporal two-stage diffusion model. \textbf{Pose Representation:} We introduce a compact representation that models hands in parametric space. To couple the hands to the object, we represent the global hand positions relative to the normalized object position at the initial frame, and further include distance information between the 3D hand joints and the object surface. While encoding distances helps reason about local shape, representing coordinates relative to the initial frame allow us to better adjust to unseen object geometry compared to per-frame object-relative pose representation used in existing work~\cite{D-Grasp}.
\textbf{Controllability:} We provide two ways to add more control to our diffusion models' output. First, we propose control through fine-grained textual descriptions. To this end, we contribute textual descriptions to the GRAB dataset \cite{GRAB} and show that they enable more robustness to unseen test prompts and increase controllability, e.g., by allowing to prompt the interacting hand and the action to perform. 
\update{Second, when a target grasp reference pose for a given object is available, we propose to leverage it to connect the grasping and interaction phases by guiding our two diffusion models at inference time.} 
The grasp reference provides a prior about how and with which hand an object should be grasped. It can either be obtained from from an image-based pose estimator~\cite{pavlakos2024reconstructing} or grasp synthesis \cite{zhang2024graspxl} as we show in our experiments in \secref{exp:qual}. 

To demonstrate the effectiveness of our approach, we run perceptual studies and experiments on the GRAB dataset \cite{GRAB}. We first compare against IMoS \cite{IMoS}. Our results indicate that our method generates better results across a variety of physics-based and motion metrics. To justify our technical contributions, we compare our final model against human-body diffusion baselines by adapting these methods to HOI. We show that we generate motions of higher quality while enabling more control of the outputs reaching feasible poses. Furthermore, we demonstrate that detailed textual descriptions increase robustness to unseen texts and offer better control of the diffusion model than training with heuristics-generated text descriptions. Decoupling our method into two stages and grasp guidance brings better generalization and high practical value to \methodname~through controllability, e.g., for downstream tasks such as animation or generating synthetic data at scale. We show this with two applications for grasp guidance: i) using an image-based pose estimator~\cite{pavlakos2024reconstructing} and ii) leveraging pre-generated grasp poses~\cite{zhang2024graspxl} to define target reference grasps, and consequently sampling multiple different actions through text in the interaction stage. 

We contribute the following: \textbf{(i)} To the best of our knowledge, DiffH2O is the first method that synthesizes hand-object interactions on unseen objects from textual descriptions. \textbf{(ii)} A temporal two-stage diffusion process that splits generation into grasping and action-based interaction, which improves generalization to different textual prompts. \textbf{(iii)} Grasp guidance and subsequence imputing that can be applied at inference to the diffusion process to increase the controllability of  outputs and improve generalization to unseen objects. \textbf{(iv)} Detailed textual descriptions to the GRAB dataset. Code and data will be made public upon acceptance\updatered{\footnote{\updatered{For code and data, see our project website here: \url{https://diffh2o.github.io}}}}.

\section{Related Work}
\label{related}

\subsection{Hand-Object Synthesis}
Recent efforts in interaction synthesis have largely been driven by the surge of high-quality human-object interaction datasets \update{\cite{GRAB, brahmbhatt2019contactdb, hampali2020ho3d, brahmbhatt2020contactpose, chao2021dexycb, kwon2021h2o, fan2023arctic, liu2022ho4d, hasson19_obman, YangCVPR2022OakInk, zhan2024oakink2, wang2023dexgraspnet}}. Some studies focus on generating coarse full-body object interactions \cite{zhang2022couch,wang2021scene,lee2023locomotionactionmanipulation, Hassan:SIG:2023, Dynamics_regulated}, such as carrying boxes. FLEX \cite{tendulkar2023flex} trains a hand and body pose prior and optimizes the priors to achieve diverse, static full-body grasps. GOAL \cite{GOAL} and SAGA \cite{SAGA}  use CVAEs to generate approaching motions for full-body grasps, whereas TOHO \cite{li2024task} models both approaching and manipulation tasks via neural implicit representations.  Similarly, \citet{braun2023physically} model the full range of motion, leveraging physics simulation and reinforcement learning (RL). In contrast, we focus on fine-grained HOI and generalization to unseen objects.
Among methods that model full-body object interaction, closest to our work is IMoS \cite{IMoS}, a two-stage method to generate HOI on seen objects based on language prompts. Starting from a grasping state, they first generate body motions and then optimize for object trajectories using a heuristics-based optimization to model HOI. By contrast, our model directly predicts the hand and object poses, models both approaching and manipulation, and generalizes to unseen objects.  

Another line of research focuses on generation of hand-object interaction sequences in isolation from full-body due to wide applications in VR and the need for a dedicated model for fine-grained details of finger motion. One prominent solution is to turn to physical simulation and RL, e.g., by learning dexterous manipulation tasks in simulation from full human demonstration data either collected via teleoperation~\cite{rajeswaran2017rss} or from videos~\cite{garciahernando2020iros, qin2021dexmv}. \citet{mandikal2020graff} propose a reward function that incentivizes dexterous robotic hand policies to grasp in the affordance region of objects. \update{\citet{li2024grasp} propose a solution for grasping multiple objects at once.} \citet{she2022grasping} propose a new dynamic state representation to generate dexterous grasps\update{, whereas other works leverage implicit representation \cite{karunratanakul2020grasping} or force-closure \cite{liu2021synthesizing}.} D-Grasp~\cite{D-Grasp} \update{and Unidexgrasp~\cite{xu2023unidexgrasp}} leverage RL and physics simulation to generate diverse hand-object interactions from sparse reference inputs. ArtiGrasp~\cite{zhang2024artigrasp} extends this to two-handed grasps and generates articulated object motions. \update{Unidexgrasp++\cite{wan2023unidexgrasp} proposes a solution that transfers to vision-based inputs, which may be used for robotic grasping. These works focus on achieving stable grasping, whereas we model both the grasping and manipulation of objects.}  
 
To model HOI, another solution is to rely on purely data-driven frameworks. For example, \cite{zhou2022toch, anonymous2023geneoh, taheri2024grip} propose methods that enable denoising hand-poses from noisy sequences of hand-object poses. \citet{ye2012synthesis} predict the local hand pose given full body and object motion. Similarly, ManipNet~\cite{Manipnet} predicts local hand poses for two-handed interactions based on 
wrist-object trajectories. Given object motion of articulated objects, CAMS~\cite{zheng2023cams} predicts one-handed poses that align with the object motion, whereas \citet{chen2023nonprehensile} focus on grasping objects using dexterity in the environment. All 
these works either assume a 
hand-object trajectory, focus only on one hand, ignore semantic action information, or do not generalize well to unseen objects. In contrast, we propose a framework that allows the synthesis of two-handed object interactions from text as well as generalization to unseen objects.

\subsection{Diffusion for Motion Synthesis}
Diffusion models \cite{sohl2015deep} have gained popularity in many domains such as image generation \cite{ho2020denoising}, video generation \cite{yang2023diffusion}, audio synthesis \cite{kong2021diffwave} or hand reconstruction~\cite{ye2023affordance,ye2023vhoi}. They have also been adapted to human motion synthesis \cite{tevet2023mdm,zhang2022motiondiffuse}. Various improvements have been proposed by integrating physics \cite{yuan2023physdiff}, scene-awareness \cite{huang2023diffusionbased}, increasing efficiency by diffusing in a pre-trained latent space \cite{chen2023mld}, or the composition of multiple actions \cite{athanasiou2022teach}. To enable more controllability during motion synthesis, \citet{karunratanakul2023gmd} propose GMD that guides the diffusion model towards target objectives at inference time. However, GMD contains design choices specific to human motion generation, by first generating a 2D root trajectory over the whole sequence, and then generating corresponding full-body poses, which is not directly transferable to HOI.
While these works focus on human motion in isolation from objects,
InterDiff~\cite{xu2023interdiff} generates human-object interactions via diffusion models. Their work is improved upon by recent concurrent works \cite{diller2023cghoi, li2023hoi, peng2023hoi} that leverage contact-based predictions in combination with inference time-guidance to improve the interaction quality. However, these works focus on full-body motions and neglect intricate hand-object interactions. In contrast, we focus on the detailed interactions of fingers and objects. Most similar to ours, the concurrent work MACS \cite{shimada2023macs} proposes a diffusion model for hand-object motion synthesis. 
Their focus is on synthesizing interactions for a single object with varying mass, while our focus is on generalization to unseen shapes and interaction synthesis conditioned on text. 








\begin{figure*}[ht!]
\resizebox{0.8\textwidth}{!}{
\includegraphics[width=1.0\textwidth]
{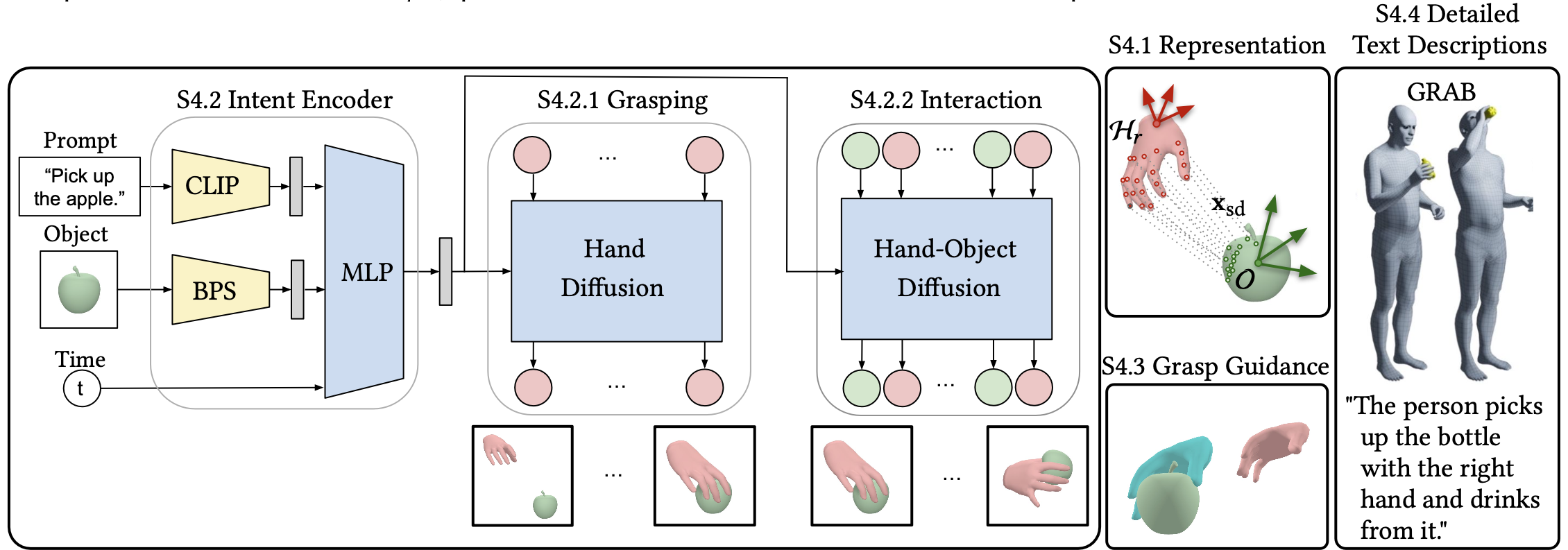}
}

\caption{\textbf{Overview of \methodname.} We couple hands and objects by representing hands relative to the object position in the initial frame and encoding hand-object distances (Sec.~\ref{sec:representation}). We observe that objects are static until they have been grasped, and propose to decouple grasping and interaction stages and modelling them with two different diffusion processes (Sec.~\ref{sec:prepost}). Finally, we make use of grasp guidance and subsequence imputation to ensure a smooth transition between these two stages (Sec.~\ref{sec:keyframe_guidance}). We further show fine-grained synthesis controllability through our detailed textual descriptions (Sec.~\ref{meth:tex_aug}).}
\label{fig:architecture}
\end{figure*}

\section{Preliminary: Diffusion Models}
\label{sec:diffusion_models}

Denoising Diffusion Probabilistic Models (DDPM)~\cite{sohl2015deep} model the probability distribution of a given dataset for images~\cite{stablediffusion}, videos~\cite{ho2022video} or motion sequences~\cite{tevet2023mdm}. 
The diffusion model involves a forward process, which is a Markov chain consisting of sequentially adding Gaussian noise to the data, and a reverse process to denoise the data gradually. The forward process results in a distribution,
$q(\textbf{x}_t | \textbf{x}_{t-1})$,
where \update{a clean input $\textbf{x}_0$ is gradually noised to $\textbf{x}_T$, with  $t\in [0, T]$ as the total number of diffusion steps.} The reverse process models the probability distribution, $p_\theta(\textbf{x}_{t-1}\lvert \textbf{x}_{t})$, \update{where the noisy input $\textbf{x}_T$ is denoised into $\textbf{x}_0$}, which is modeled via a neural network with parameters $\theta$. The process of data generation involves sequential denoising and noising steps. The denoising is modeled in auto-encoder fashion predicting either directly the denoised output, $\textbf{x}_{0,\theta}(\textbf{x}_t, t)$; the mean, $\bm{\mu}_\theta(\textbf{x}_t,t)$, or the noise, $\bm{\epsilon}_\theta(\bold{x}_t,t)$, used in the noising process. The auto-encoder is trained to minimize
\begin{equation}
    \mathcal{L}_{\text{Diff}} = \mathbb{E}_{\epsilon \sim \mathcal{N}(0,1), t} \mathcal{L}_{ae} ,    
\end{equation}
where $\mathcal{L}_{ae}$ is either $\lvert\lvert \textbf{x}_{0,\theta}(\textbf{x}_t, t) - \textbf{x}_0\lvert\lvert$, $\lvert\lvert \bm{\mu}_{\theta}(\textbf{x}_t, t) - \bm{\mu}\lvert\lvert$, or  $\lvert\lvert \bm{\epsilon}_\theta(\bold{x}_t,t) - \bm{\epsilon} \lvert\lvert^2_2$. \update{Depending on the application, a different choice of the model output works better in practice, e.g., directly predicting the output $\textbf{x}_{0}$ for human motion \cite{tevet2023mdm} or noise prediction $\bm{\epsilon}$ for images \cite{stablediffusion} or guidance \cite{karunratanakul2023gmd}.} Conditional diffusion models generate output conditioned on various kinds of input such as text~\cite{stablediffusion} or other semantic information~\cite{controlnet}. Such a model is obtained by providing the condition as additional signal, $\textbf{c}$, to the auto-encoder, resulting in outputs $\textbf{x}_{0,\theta}(\textbf{x}_t, t, \textbf{c})$, $\bm{\mu}_\theta(\bold{x}_t,t,\textbf{c})$, or $\bm{\epsilon}_\theta(\bold{x}_t,t,\textbf{c})$.

Diffusion models can be guided to generate outputs of the required form. Two popular methods to guide the diffusion models are classifier guidance~\cite{dhariwal2021diffusion} and classifier-free guidance~\cite{ho2022classifier}. We describe the classifier-guidance method here and refer the reader to \cite{ho2022classifier} for details on classifier-free guidance.
The classifier-guidance method approximates $p_\theta(\textbf{x}_{t-1}\lvert\textbf{x}_t,\textbf{c})\propto p_\theta(\textbf{x}_{t-1}\lvert\textbf{x}_t) p(\textbf{c}\lvert\textbf{x}_{t})$, where $\textbf{c}$ is the guidance signal. Deriving from this, \citet{dhariwal2021diffusion} show that the new sampling process can be expressed as 
\begin{equation}
\bm{\mu}_\theta(\textbf{x}_t,t) = \bm{\mu}_\theta(\textbf{x}_t,t)' + s\beta\nabla_{\textbf{x}_t}\log{p(\textbf{c}\lvert\textbf{x}_t)},    \label{eq:guided_diff_rev} 
\end{equation}
where $\bm{\mu}_\theta(\textbf{x}_t,t)'$ indicates the original mean, $s$ is the scaling of the gradients, and $\beta$ is a variance scheduler. Here the mean is shifted by the scaled gradient. Intuitively, the scaled gradient of the guidance signal, $\nabla_{\textbf{x}_t}\log{p(\textbf{c}\lvert\textbf{x}_t)}$ nudges the denoised mean $\bm{\mu}_\theta$ to generate samples in accordance with the guidance $\textbf{c}$. $\log{p(\textbf{c}\lvert\textbf{x}_t)}$ could represent the class probability which can be implemented using a neural-net, or could represent a cost term $G(\textbf{x}_t)$ that is optimized. The classifier guidance method has the advantage that it can use pre-trained diffusion models without retraining them and can benefit from any guidance signal that is differentiable to guide the diffusion.

\section{Method}
\label{sec:method}
We focus on the task of generating a sequence of hand and object poses $ \bold{x} \coloneqq (\mathcal{H}_l, \mathcal{H}_r, \mathcal{O} )$, where $(\mathcal{H}_l, \mathcal{H}_r)$ indicate a sequence of left and right hand poses, and $\mathcal{O}$ is a sequence of object poses given some input conditioning signal $\textbf{c}$, which in our case comprises a text description and the object mesh. In other words, we want to model a conditional probability distribution $p(\bold{x} \, | \, \textbf{c}, G=0)$ with a motion DDPM. The optional term $G=0$ \update{can be any} differentiable goal function that should be minimized, e.g., the distance to a target grasp \update{in our case (see Sec. \ref{sec:guidance})}. See \figref{fig:architecture} for an overview.
In Sec.~\ref{sec:representation}, we introduce our compact representation that couples the hand and object poses. We propose a two-stage generation of hand-object interactions in Sec.~\ref{sec:prepost} and grasp guidance in Sec.~\ref{sec:guidance}. Lastly, in Sec.~\ref{meth:tex_aug}, we introduce our newly-collected detailed textual annotations for GRAB that enable more controlled generation of HOI.

\subsection{Canonicalized Hand-Object Representation}
\label{sec:representation}

Our proposed canonicalized representation \update{$\mathbf{x} = (\mathcal{O}, \mathcal{H}_{l}, \mathcal{H}_{r}) \in \R^N$, where $N$ is the number of steps in the sequence,} contains information about both hands and an object. The object poses are defined as $\mathcal{O}= (\bm{\tau}_o, \bm{\phi}_o)$, where $\bm{\tau}_o \coloneqq (\bm{\tau}_o^0, \bm{\tau}_o^1, \dots , \bm{\tau}_o^\text{N-1})$ is a sequence of 3D object locations, with $\bm{\tau}^i_o \in \R^3$, and $\bm{\phi}_o \coloneqq (\bm{\phi}_o^0, \bm{\phi}_o^1, \dots , \bm{\phi}_o^\text{N-1})$ is a sequence of 3D object orientations, with $\bm{\phi}_o^i \in \R^6$ in 6D representation \cite{zhou2019continuity}. The hands are represented by:
$\mathcal{H}_{j}= (\bm{\tau}, \bm{\phi}, \bm{\theta}, \mathbf{x}_{\text{sd}}), j \in \{ l, r \}$, where $l$ denotes the left and $r$ the right hand, respectively.
This representation is based on the parametric MANO hand model \cite{MANO}. Specifically, the global 3D positions of the hands are given by $\bm{\tau} \in \R^{3xN}$ and the global 3D orientations are represented by $\bm{\phi} \in \R^{6xN}$. To achieve more natural poses, we define the local hand pose in PCA space of MANO with the first 24 components as $\bm{\theta} \in \R^{24\update{xN}}$. We further encode signed distances, $\mathbf{x}_{\text{sd}} \in R^{21\text{x}3\update{\text{xN}}}$, between each hand joint and its closest point on the object mesh. As we show in ablations (see \tabref{tab:pose_repr_ablation}), \update{this approximation of a signed-distance field (SDF),} coupled with our diffusion model reduces physical artifacts. 
Existing works \cite{D-Grasp} encode hand-object poses with a per-frame object-relative representation, which tightly couples the hands to the training object pose and therefore limits generalization to unseen objects. In contrast, we represent hand positions relative to the normalized object position \updatered{at the frame that marks the boundary between the grasp and interaction stage}. This representation is less variant to specific training object pose and more correlated with the global poses involved in HOI, which is often shared across different training objects in HOI datasets \cite{chao2021dexycb, GRAB}. As a result, it yields better generalization and accuracy for HOI across different (unseen) objects, as we show in ablations in Sec.~\ref{exp:ablations}.

\subsection{Two-Stage Hand-Object Generation}
\label{sec:prepost}
We split the synthesis into grasping and interaction stages to facilitate generalization to unseen objects. The grasping stage models the motion from an initial pose to object grasp. The interaction stage performs an action-based manipulation of the object. Our two-phase approach is motivated by the fact that any hand-object interaction involves an approaching phase where one or two hands grasp a static object, followed by an intent-driven interaction with the object. By decoupling the action-based manipulation from grasping, we can leverage the entire set of motions in the data for training the grasping phase, irrespective of the action that is performed in the interaction phase. These insights help counteract the limited scale of the data and improve generalization to different textual inputs.

As shown in Fig.~\ref{fig:architecture}, our grasping and interaction diffusion models are conditioned on three types of inputs, namely, 1) text prompts describing the object and the action to be carried out, 2) object mesh which provides the geometry of the object, and 3) the time step $t$ in the denoising process. We use CLIP~\cite{radford2021learning} embeddings to encode the text prompt denoted by $\mathcal{T}$, BPS~\cite{prokudin2019efficient} to encode the object mesh denoted by $\mathcal{M}$ and an MLP to encode the time step as in previous works~\cite{ dhariwal2021diffusion}. We provide more implementation details in supplementary material and describe the grasping and interaction phases below in detail.

\subsubsection{Grasping Stage}
\label{sec:pre}
This stage is defined as a sequence containing one or two hands approaching a static object from a rest pose until the object is grasped. As the object is static throughout this phase we do not predict the object motion and only model the hand pose sequence. To obtain the training data for this phase we use heuristics to determine sequence boundaries from the larger action sequences. \update{In particular, the boundary is defined as the first frame where the object's lateral and vertical velocity surpass a threshold of 0.01m/s and at least 7 vertices of the hand are in contact with the object . We then} normalize the sequence lengths by interpolating or downsampling the poses. The text description in this phase is a generic sentence ("The person grasps the <object>.").

We use the $\epsilon_\theta$ diffusion model that directly predicts the noise in classifier-free manner~\cite{ho2022classifier} which is more suitable for inference-time guidance~\cite{karunratanakul2023gmd}, e.g., grasp guidance (Section~\ref{sec:keyframe_guidance}). In other words, our grasping diffusion model training loss is given by 
\begin{equation}
\mathbb{E}_{\epsilon \sim \mathcal{N}(0,1),t}\lvert\lvert \epsilon_\theta(\bold{x}_t^g,\mathcal{T}, \mathcal{M}, t) - \epsilon \lvert\lvert^2_2,     
\end{equation}
where $\bold{x}_t^g$ is the synthesized motion at diffusion step $t$ and the final output of the grasping stage after denoising  $\bold{x}_t^g$ with $\epsilon_\theta$ is $\bold{x}_0^g$.

\subsubsection{Interaction Stage}
\label{sec:post}
This stage comprises the motion of both hands and the object after a grasp has been established. The hands manipulate and interact with the object according to the action defined through the textual prompt. We use the same sequence boundary determined in the grasping stage to get the training data for interaction. We employ a $\textbf{x}_0$ diffusion model in a classifier-free manner (see \secref{sec:diffusion_models}), i.e., we directly predict the denoised output $\textbf{x}_0^i$, that  
result in less noisy, higher quality motion than $\epsilon_\theta$ diffusion model for the interaction phase that does not involve inference-time grasp guidance. 
 Our training loss is given by
\begin{equation}
    \mathbb{E}_{\epsilon \sim \mathcal{N}(0,1),t}\lvert\lvert \textbf{x}_{0,\theta}^i(\bold{x}_t^i,\mathcal{T}, \mathcal{M}, t) - \textbf{x}_0^i \lvert\lvert^2_2.
\end{equation}

\subsubsection{Grasp to Interaction Stage Transition}
\label{sec:keyframe_guidance}
Maintaining a seamless transition between the grasping and interaction outputs is critical to generating realistic motion sequences. We utilize motion imputing on the interaction model to achieve a smooth transition. In particular, we impute the start of the interaction sequence from the entire sequence obtained by the (guided) grasping model. Intuitively, this adjusts the generative process for interaction stage based on the observed grasping motion. To perform the imputation, we define the projection $P_{g}^i$ that resizes the hand poses $\textbf{x}_0^{g} \in (\mathcal{H}_l, \mathcal{H}_r)$ from the grasping stage to hand-object poses $\bold{x}_0^i$ from the interaction stage by filling in zeros, both temporally and spatially. Similarly, $M_{g}^i$ represents the imputation regions of the grasp sequence on $\bold{x}_{0}^i$. At each denoising step, the imputed denoised output $\Tilde{x}_{0,\theta}^i$ is given by
\begin{equation}
    \Tilde{\bold{x}}_{0,\theta}^i = (1-M_{g}^i) \odot \bold{x}_{0,\theta}^i + M_{g}^i \odot P_{g}^i \textbf{x}_{0}^g,
\end{equation}
where $\odot$ is elementwise multiplication. Our proposed technique, dubbed \emph{subsequence imputing}, leads to smoother transition between grasp $\&$ interaction compared to single frame transition (see Sec.~\ref{exp:ablations}).


\subsection{Grasp Guidance}  
\label{sec:guidance}
To add controllability to the generation and improve  generalization to unseen objects, we introduce grasp guidance for the grasping stage, which can be optionally added at inference time \update{via the goal function $G$ (see Sec. \ref{sec:diffusion_models} and Eq. \ref{eq:guided_diff_rev})}. We leverage a single grasp frame $\hat{\textbf{h}}^g_0$ as target grasp (hand pose) for the last frame of the grasping stage. It provides a prior about how and with which hand an object should be grasped. Such a grasp can be obtained from grasp generation \cite{zhang2024graspxl} or from an off-the-shelf hand-object pose estimator~\cite{pavlakos2024reconstructing} as demonstrated in Sec.~\ref{exp:qual}.
We aim to minimize the goal function $G(\textbf{x}^g_t)=0$ and approximate the gradient of the guidance signal in Eq.~\ref{eq:guided_diff_rev} as
\begin{equation}
    \nabla_{\textbf{x}_t^g} \log{p(\textbf{c}\lvert \textbf{x}_t^g)} = -\nabla_{\textbf{x}_t^g} G(\textbf{x}^g_t)  \approx -\nabla_{\textbf{x}_t^g}\lvert \lvert \textbf{h}^g_{0,\theta}(\textbf{x}_t^g) - \hat{\textbf{h}}_{0}^g \lvert\lvert_2^2,
    \label{eq:guidance_grad}
\end{equation}
where $\textbf{h}_{0,\theta}^g \in (\mathcal{H}_l, \mathcal{H}_r)$ is the hand(s) pose in the last frame of the grasping sequence after denoising $\textbf{x}_t^g$ \update{and the cost term $G(\textbf{x}^g_t)$ minimizes the distance of the hand(s) pose to the target grasp}. \update{Depending on the availability, the guidance signal may include one or two hands}. Since the diffusion process is a denoising model of a motion sequence, it relates individual frames to other frames in the sequence. Hence, by backpropagating through the diffusion model (via autodiff), we can compute a dense guidance signal for all frames, even if only sparse guidance signals (one in our case) are provided.

\subsection{Detailed Text Descriptions}
\label{meth:tex_aug}
Our model generates hand-object interactions based on textual inputs. To the best of our knowledge, there is currently no HOI dataset that contains detailed textual descriptions. 
For instance, the GRAB dataset~\cite{GRAB} only provides categorical action labels.  To address this, we auto-generated sentences using the template ``the person <verb> + <object>’’~\cite{IMoS} (referred to as ``simple’’ text). However, these sentences lack detailed information, which limits the controllability of the model via textual inputs. Therefore, we contribute carefully annotated textual descriptions of GRAB (referred to as ``detailed" sentences). 
We instructed our annotators to carefully watch each ground truth video and describe the actions in three distinct stages of \emph{pre-action}, \emph{action}, and \emph{post-action}. These descriptions are further augmented with  hand and positional information. An example from the dataset is as follows: "The person picks up the apple with the right hand, passes it to their left hand, and places it back with their right hand”. These detailed descriptions enhance the dataset, enabling more accurate generation and precise control of HOI as we show in \refsec{exp:text_aug}.

\begin{table*}[t!]
  \centering
    \caption{\textbf{Comparison to State-of-the-Art in the Interaction Stage.} We compare our method with two backbones (transformer and UNet) against IMoS~\cite{IMoS}. We  report results on an unseen subject split (top 3 rows)~\cite{IMoS}, and on our unseen object test dataset (bottom 3 rows). For IMoS, we use the same pretrained model, trained on unseen subject split, across all our experiments (unseen subject/object splits), due to difficulties in reproducing training performance for the unseen object split (indicated with  IMoS* in the table). $\uparrow$: higher values are better, $\downarrow$: lower values are better.}
    \resizebox{0.9\textwidth}{!}{%
  \begin{tabular}{@{}l|c|c|cccccc@{}}
    \toprule
      &Method & Backbone  & SD [$m$] ($\uparrow$) & OD [$m$]($\uparrow$) & IV [$cm^3$] ($\downarrow$) & ID [$mm$] ($\downarrow$) & CR $(\uparrow)$ 
      & \begin{tabular}[c]{@{}c@{}} AR ($\uparrow$)\end{tabular}\\ \midrule\midrule
    \parbox[t]{11mm}{\multirow{3}{*}{\rotatebox[origin=c]{90}{ \begin{tabular}[c]{@{}c@{}} \hspace{-1.8mm} Unseen\\ \hspace{-2mm} subject \\ \hspace{-2mm} split \end{tabular}}}} 
    & IMoS~\cite{IMoS} & CVAE & 0.002 & 0.149  & 7.14 & 11.47 & 0.05 
    & 0.588 \\
    & DiffH$_{2}$O & Transformer & 0.088 & 0.185 & 6.65 & 8.39 & \textbf{0.067} & 
    0.810 \\
    & DiffH$_{2}$O & UNet & \textbf{0.109} & \textbf{0.188} & \textbf{6.02} & \textbf{7.92} & 0.064 & \textbf{0.875} \\ 
    \midrule
    \parbox[t]{11mm}{\multirow{3}{*}{\rotatebox[origin=c]{90}{\begin{tabular}[c]{@{}c@{}} \hspace{-1.3mm}Unseen\\ \hspace{-2mm} object \\ \hspace{-2mm} split\end{tabular}}}} 
    & IMoS*~\cite{IMoS}  & CVAE & 0.002 & 0.132  & 10.38 & 12.45 & 0.048 & 
    0.581\\
& DiffH$_{2}$O & Transformer & 0.133 & \textbf{0.185} & \textbf{7.99} & \textbf{10.87} & 0.073 & 
0.803 \\
    & DiffH$_{2}$O  & UNet & \textbf{0.134} & 0.179  & 9.03 & 11.39 & \textbf{0.086} & 
    \textbf{0.837} \\
    \bottomrule
  \end{tabular}
  }

  \label{tab:motion_synthesis}
\end{table*}

\section{Experiments and Results}
\label{sec:experiment}


\subsection{Baselines}

Our method focuses on synthesizing detailed hand-object motion based on text input. To the best of our knowledge, DiffH$_2$O is the first method to tackle this problem. While there is no direct baseline for hand-object motion synthesis based on language, the closest to our approach is IMoS~\cite{IMoS}, which focuses on text-based whole-body human-object interaction synthesis. IMoS first generates a body articulation based on an instruction label (action + object), and then optimizes for the object pose using a grasp heuristic. Since our paper focuses on hand articulations and object motions, we omit the full-body movements from our evaluations.

To further compare~\methodname~against human-body diffusion models, such as~\cite{tevet2023mdm} and~\cite{karunratanakul2023gmd}, we adopted them to the HOI setting. All variants use our proposed pose representation. MDM \cite{tevet2023mdm} does not use guidance, and for GMD \cite{karunratanakul2023gmd}, we use gradients to guide the model towards the grasp frame, but omit the 2D-trajectory generation stage because it is specific to human motions. 

\subsection{Data}
\subsubsection{GRAB} We utilize the subject-based split of the GRAB dataset \cite{GRAB}, \update{which contains 1335 sequences}, proposed in IMoS to run a direct comparison. However, as this split does not contain unseen objects, we also create a new split featuring unseen objects based on object similarity and semantics (see supplemental material). To assess our model's ability to generalize to unseen classes, we introduce a new split comprising unseen objects that were not present in training. \updategreen{{Our train and test split contains 1125 and 210 sequences, respectively, which we mirror to a total of 2250 and 420 sequences. To split the GRAB sequences into the grasping and interaction stages for training, we apply the heuristics described in \secref{sec:pre}} Our test set is composed of the following objects: ``apple, mug, train, elephant, alarm clock, small pyramid, medium cylinder and large torus''. Note that the training set includes pyramids, cylinders, and torus in other sizes. We exclude small pyramids, medium cylinders, and large torus from the training set to test our models' ability to generalize to different sizes of the same objects} \update{To evaluate grasp guidance in a controlled manner, we utilize the frame of the unseen test split that indicates the transition between the grasp and interaction stage. Note that the grasp reference contains both hands in this case, even if one hand is not grasping the object.}

\subsubsection{HO-3D} To further evaluate guidance and object generalization, we estimate  poses on the HO-3D dataset \cite{hampali2020ho3d} and guide our model to an estimated target hand pose.

\subsection{Evaluation Metrics}
\label{sec:metrics}
We report several physics-based metrics and  motion diversity metrics following previous works~\cite{IMoS,tendulkar2023flex,GOAL,braun2023physically}. Please see supplementary material for additional metrics based on action features. 

\subsubsection{Physics Based Metrics}
We report the interpenetration volume (IV) as the number of MANO vertices penetrating the object mesh and the maximum interpenetration depth (ID). Additionally, we compute the contact ratio (CR) as the average ratio of hand vertices within 5mm of the object mesh.

\subsubsection{Motion Metrics}
We report the sample diversity~(SD) between wrist trajectories in motion space. We sample the same input conditions 5 times, compute the pairwise Euclidean distance between the samples and report the mean SD over all test prompts. We also provide the overall diversity~(OD), which measures the pairwise Euclidean distance between all test samples.
Furthermore, we report the action recognition accuracy (AR) as in~\cite{IMoS}.

\subsubsection{Guidance Metrics}
To assess how well the models adhere to the guidance signal, we report the mean error between the predictions and the reference grasp~(GE), and the ratio at which the diffusion model maintains the correct handedness according to the reference~(HA). Furthermore, we evaluate the wrist velocities at the transition between grasping and interaction phase ($\text{T}_\text{vel}$).

\begin{table}[t!]
\centering
\caption{\textbf{Comparison to Diffusion Baselines for the Full Sequence.} We compare against HOI-adapted versions of MDM and GMD. We measure the grasp error (GE) and the accuracy of handedness (HA) with respect to the reference grasp, the interpenetration volume (IV), and  the wrist joint velocities ($\text{T}_{\text{vel}}$) at the transition between grasping and interaction.}
\resizebox{0.9\linewidth}{!}{
\begin{tabular}{c|cccc}
\hline
Method & IV [$cm^3$] ($\downarrow$) & GE $[m] \downarrow$ & HA $\uparrow$ &  $\text{T}_{\text{vel}}$ $[\frac{m}{s}]\downarrow$ \\ \hline
MDM & 8.98 & 0.38 & 0.45 & \textbf{0.14} \\
GMD & 7.47 & 0.30  & 0.58  & 0.20 \\
Ours & \textbf{7.40} & \textbf{0.12} & \textbf{0.87}  & 0.23 \\ \hline
\end{tabular}
}



\label{tab:guidance}
\end{table}

\subsection{Comparison to the State-of-the-Art}
\label{exp:main_eval}
\begin{figure*}[h!]
    \centering
    \includegraphics[width=0.95\textwidth]{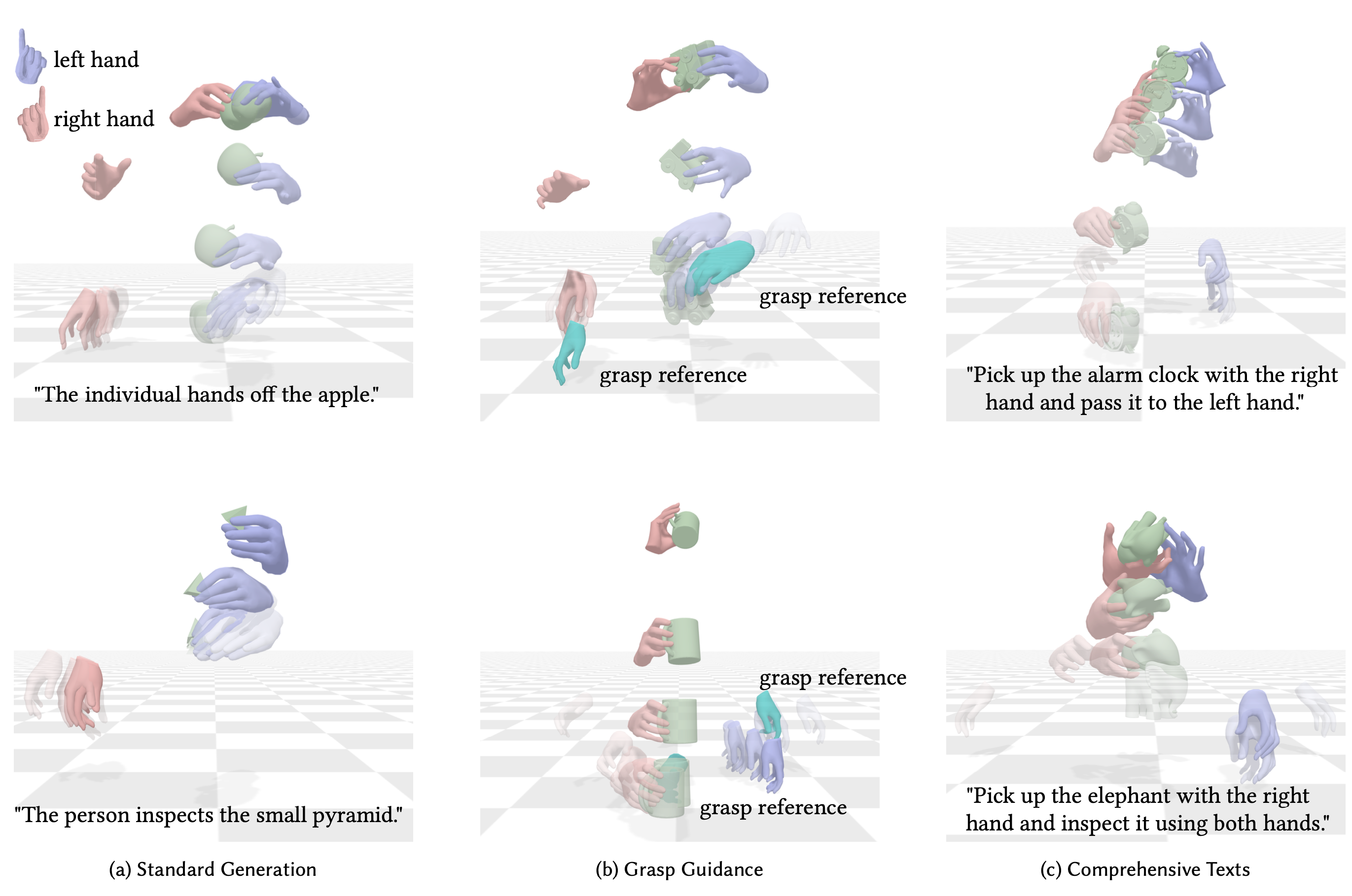}

    \caption{\textbf{Qualitative Examples}. We provide more qualitative examples with a) standard generation without any guidance b) grasp guidance c) our model trained with detailed text descriptions.}
    \label{fig:more_results}
\end{figure*}

\begin{figure*}[h!]
    \centering
    \begin{subfigure}[b]{0.25\textwidth}
        \centering        \includegraphics[width=\textwidth]{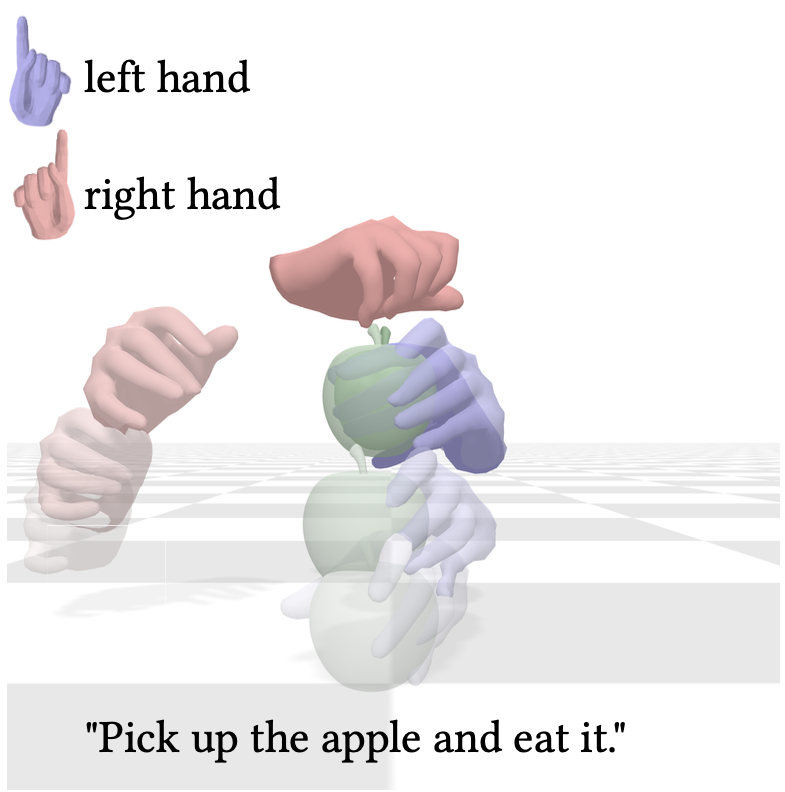}
        \caption{Wrong action}
        \label{fig:fail_sub1}
    \end{subfigure}
    \hfill 
    \begin{subfigure}[b]{0.25\textwidth}
        \centering
        \includegraphics[width=\textwidth]{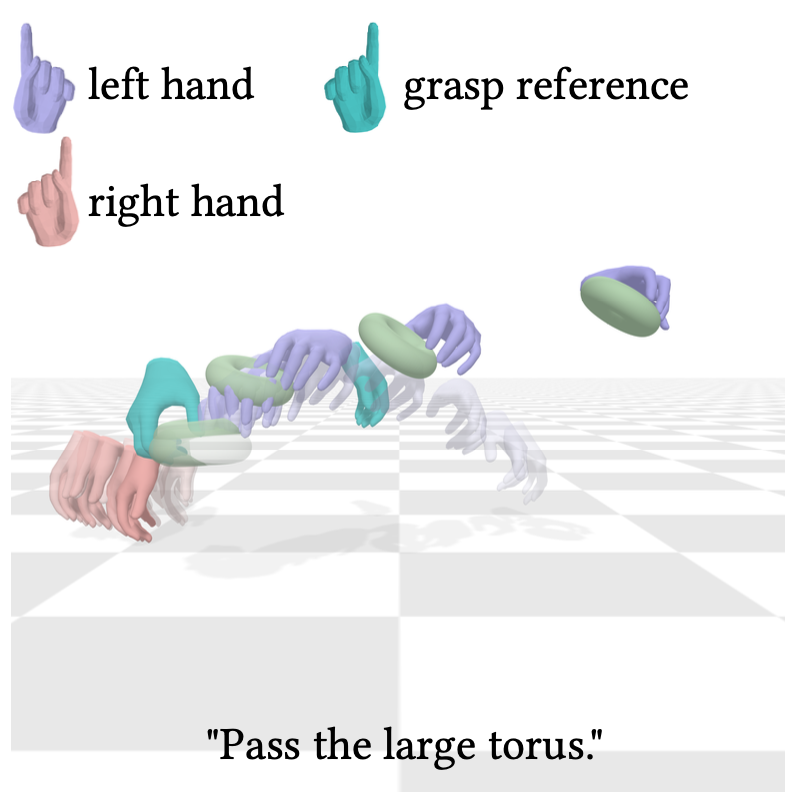}
        \caption{grasp guidance failure}
        \label{fig:fail_sub2}
    \end{subfigure}
    \hfill 
    \begin{subfigure}[b]{0.25\textwidth}
        \centering
    \includegraphics[width=\textwidth]{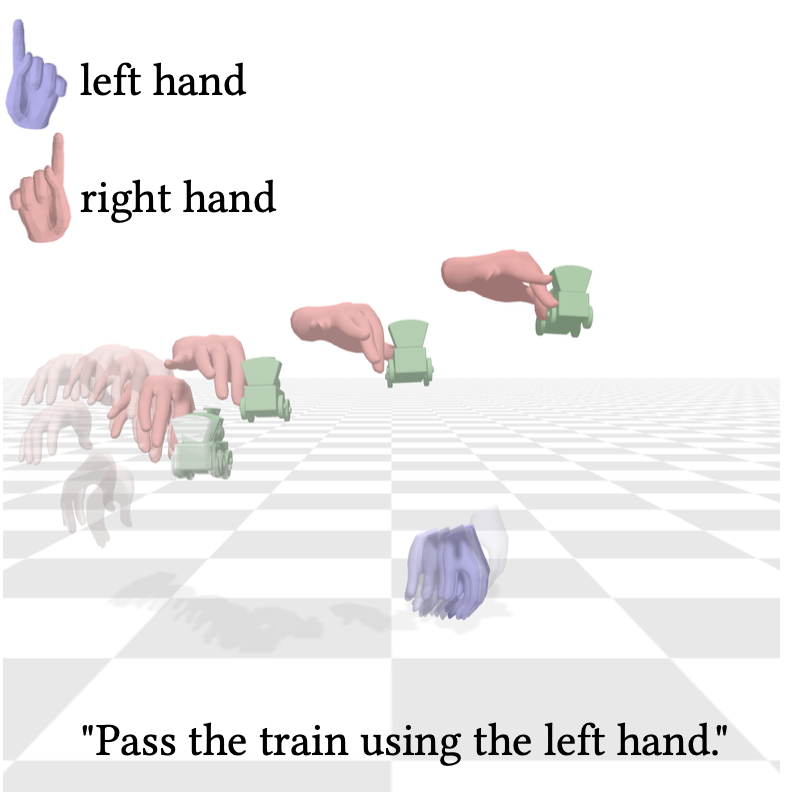}
        \caption{Handedness failure}
        \label{fig:fail_sub3}
    \end{subfigure}
    \caption{\textbf{Failure Cases}. We present three possible failure cases of our method. a) The generated motion does not match the action described in the input prompt, such as trying to perform a bottle opening motion with an apple. b) During grasp guidance, the reference grasp is largely ignored in the diffusion process, resulting in an interaction that is distinct from the grasp reference. c) Despite training with our curated text annotations, the model sometimes does not pick up on the cue of handedness and may interact with a hand different from the one provided in the text prompt.}
    \label{fig:fail}
\end{figure*}

We conduct a set of experiments to compare to relevant baselines quantitatively. \updategreen{We present qualitative examples of our method in \figref{fig:more_results} and failure cases in \figref{fig:fail}.}

\subsubsection{Setting 1 - Interaction Only}
We compare our approach against the closest existing work, IMoS, in their interaction-only setting and report results in \tabref{tab:motion_synthesis}. To ensure a fair comparison with IMoS, we evaluate our model without the grasp guidance and without our detailed text annotations. We test \methodname~with two backbones, transformer as in~\cite{tevet2023mdm} and UNet as in~\cite{karunratanakul2023gmd}. Our model significantly outperforms IMoS across all physics and motion diversity metrics. In particular, our method yields motions that are significantly more diverse (higher SD and OD), exhibit less interpenetration (lower IV and ID), and align better with the action type (higher AR). Note that IMoS assumes the availability of a ground-truth test motion to initialize the first few frames, while ours predicts the entire sequence. Yet, our approach achieves considerable performance improvements. While IMoS provides compelling realistic full body synthesis results, fine-grained finger motions and object motions lag behind the realism and diversity demonstrated by our method, as shown in \figref{fig:qualitative} and in our supplementary video. 
In \secref{sec:user_study}, we further provide a perceptual study to compare our method against the baseline.

\subsubsection{Setting 2 - Grasping + Interaction}
In this experiment, we focus on generating entire sequences of grasping and interaction, evaluating how well the models can produce high-quality motions without physical artifacts and adhere to guidance signals. Results in \tabref{tab:guidance} show our method outperforms human motion diffusion baselines (MDM/GMD) with better grasp accuracy (GE), handedness (HA), and naturalness (IV). This validates our temporal two-stage design with grasp guidance and subsequence imputing over previous diffusion models for the setting of hand-object interactions.

\subsection{Perceptual User Study}
\label{sec:user_study}
\begin{figure}[t!]
\centering

\includegraphics[width=0.45\textwidth]{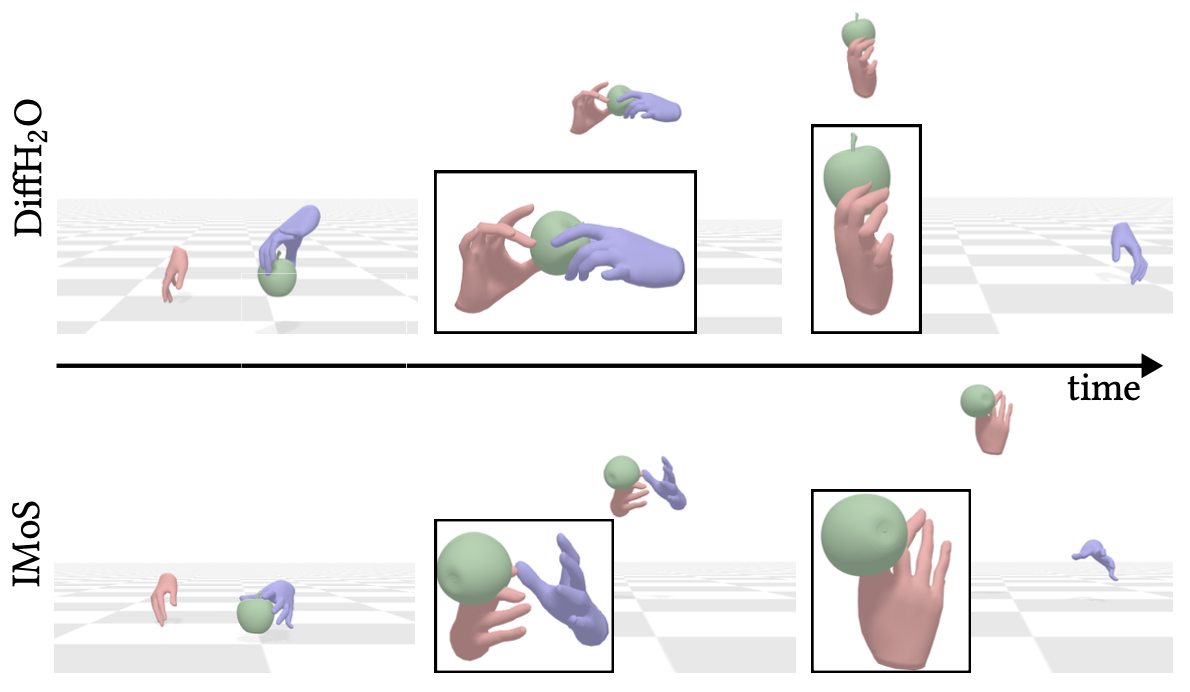}

\caption{\textbf{Qualitative Comparison.} Post-optimizing object motion as in IMoS \cite{IMoS} (bottom row) exhibits artifacts with fine-grained manipulations, e.g., when an object switches hands.  In contrast, our approach (top row) seamlessly handles such cases. Best seen in supplemental video.}
\label{fig:qualitative}
\end{figure}

\updategreen{To evaluate the visual quality of our motions, we conducted a perceptual user study on a set of 21 participants, who were randomly assigned one of four sets, each consisting of 30 randomly selected sequences. We compared our approach against IMoS. We displayed results from our method and IMoS side-by-side in random order, along with the input sentence description used to synthesize motions. We asked the following questions to the participants: ``Which sequence is more realistic?'' and ``Which sequence has more more variety in motion?''. We defined a sequence to be more realistic if the overall motion looks human-like, e.g., if the object is grasped realistically and there are less artifacts such as floating objects or interpenetration. Moreover, a sequence has more variety if the object is manipulated multiple times and in distinct ways. In 63.1$\%$ of the responses, participants selected our method as the most realistic compared to IMoS. Even more distinctively, in 72.9$\%$ of the responses, participants favored our method to contain more variety than IMoS. We obtained statistically significant Fleiss' kappa scores \cite{fleiss1971measuring} of 0.34 and 0.43 for Realism and  Diversity, respectively, at p<0.05. In our investigation, our model showed low realism but high diversity for the ``pass'' action, influenced by users perceiving ``pass'' action label in the GRAB dataset as transferring an object between hands, not to another person. In the ``offhand'' action category of GRAB, where an object is passed from one hand to another (see Figure \ref{fig:qualitative} and video 02:51), our model excelled in realism, while IMoS exhibited artifacts. This is due to the fact that IMoS determines the hand which is in contact with the object according to ground-truth, which leads to a jump of the object for the ``off-hand'' category.}

\subsection{Ablation Studies}
\label{exp:ablations}
We perform several ablations to justify our technical contributions. To evaluate the controllability gained through guidance, we compare our final method with several variants in \tabref{tab:keyframe_guidance} for the full setting that generates the whole sequence with grasping and interaction (Setting 2). We use a base model which indicates a single stage model without guidance. We then gradually add our model components: (i) 2-stage design that decouples grasping and interaction~(2-ST), (ii) grasp guidance (GG), and, (iii) subsequence imputing in the interaction phase (SI) as described in Sec.~\ref{sec:prepost}. As expected, the grasp reference is being ignored without guidance (top 2 rows in \tabref{tab:keyframe_guidance}), leading to high grasp errors and low hand correctness. Adding guidance allows for better performance in grasp error and hand accuracy, however, a high $\text{T}_\text{vel}$ indicates a sudden jump of the hands between grasping and interaction. Our subsequence imputing mitigates this and leads to the best overall score at the cost of a slight increase in grasp error and decrease in hand correctness. The IV is lowest with our final model, indicating the best generalization to unseen objects. 
To validate our representation, we further compare our approach, which normalizes hands with respect to the object position at first frame, against a number of baselines: (i) a variant inspired by D-Grasp~\cite{D-Grasp} that involves a redundant representation involving joint angles \update{in Euler space} and angular velocities, (ii) our representation without SDF, and, (iii) our representation with per-frame object relative poses. As shown in \tabref{tab:pose_repr_ablation}, our final representation achieves the best overall scores and trade-off with low interpenetration and high contact ratios. As discussed in Sec.~\ref{sec:representation}, our representation relative to initial frame is less variant with specific object shapes and more correlated with motion involved in hand-object interaction compared to per-frame object-relative representation. This effectively improves generalization for HOI across different objects. In addition, SDF feature further improves accuracy by reasoning about local shape.

The D-Grasp representation has the lowest interpenetration scores, but as indicated by the contact ratio and our qualitative results in the video, the hands are often disconnected from the objects, potentially due its  challenge in optimizing in high-dimensional, redundant feature space.  \update{Additionally, our PCA space helps reduce unnatural hand poses that occur when predicting joints in angle space directly (see supplemental video). While the PCA space is beneficial in data-sparse settings, it may hinder more fine-grained finger manipulations compared to joint angle predictions.}

\begin{table}[t!]
\centering
\caption{\textbf{Ablation Study.} We provide ablations of our components against the base model. We measure the grasp error (GE) and the accuracy of handedness (HA) with respect to the reference grasp, as well as interpenetration volume (IV). We also provide the wrist joint velocities ($\text{T}_{\text{vel}}$) for the transition from grasping to interaction. Bold is the best, underlined is the second best.}
\resizebox{0.85\linewidth}{!}{
\begin{tabular}{ccc|cccc}
\hline
\multicolumn{3}{c|}{Component} & \multicolumn{4}{c}{Metrics} \\ 
2-ST  & GG & SI & IV [$cm^3$] ($\downarrow$) & GE $[m] \downarrow$ & HA $\uparrow$ &  $\text{T}_{\text{vel}}$ $[\frac{m}{s}]\downarrow$ \\ \hline
$\times$ & $\times$ & $\times$ & \underline{7.48} & 0.31 & 0.52 & \textbf{0.21} \\
\checkmark & $\times$ & $\times$ & 9.03 & 0.49  & 0.46  & 12.68 \\
\checkmark & \checkmark & $\times$ & 8.52  & \textbf{0.06}  & \textbf{0.97} & ~5.37 \\
\checkmark & \checkmark & \checkmark & \textbf{7.40} & \underline{0.12} & \underline{0.87}  & ~\underline{0.23} \\ \hline
\end{tabular}
}

\label{tab:keyframe_guidance}
\end{table}


\subsection{Generation from Estimated and Pre-Synthesized Pose}
\label{exp:qual}
We show two qualitative applications of~\methodname~in video. First, we run an image-based pose estimator~\cite{pavlakos2024reconstructing} on single images from HO3D~\cite{hampali2020ho3d} to obtain object-relative hand poses. Second, we generate grasping poses using GraspXL \cite{zhang2024graspxl}.
We then pass these hand poses through our method as reference grasps in grasp guidance (see Sec.~\ref{sec:guidance}), together with multiple different textual descriptions. This highlights the practicality of our framework given a single grasp reference or grasp motion, to generate multiple diverse sequences. \update{See supplementary material for details about how we obtain grasp references and further discussions on their benefits and limitations.}

\subsection{Controllability via Text Prompts}
\label{exp:text_aug}

In this experiment, we train \methodname~ once with our detailed text descriptions and once with simple text prompts generated from the GRAB objects and action labels. We measure whether the active hand in the generated motion corresponds to the prompted hand to  gauge controllability.
Results in \tabref{tab:textual_augmentations} show that detailed text descriptions significantly improve control over handedness, achieving an accuracy of $86.5\%$ vs. $59.3\%$ with simple prompts. Performance drop when training on simple texts and testing on detailed texts (0.702) and vice-versa, which was similarly observed in \cite{shvetsova2023instyle} due to a style gap.
We also measure action correctness by evaluating action recognition accuracy (AR, see Sec.~\ref{sec:metrics}). Using detailed descriptions enables our model to generate motions that align more accurately with the text input, demonstrated by a 0.887 action accuracy compared to 0.516 with our model tested with simple prompts. Since the test inputs are much more diverse with the detailed textual descriptions, this furthermore demonstrates the robustness to unseen sentences. We also report the cosine similarity between the different sets of texts as a reference.

\begin{table}[t!]
\centering
\caption{\textbf{Pose Representation Evaluation.} We compare different alternative pose representations in interaction stage and demonstrate the benefits of object-relative pose representation and encoding hand-object signed distances. Bold indicates the best result, underlined is the 2$^{\text{nd}}$ best result.}
\resizebox{0.95\linewidth}{!}{
\begin{tabular}{l|ccc}
\hline
\multirow{2}{*}{Pose Representation (unseen objects)} & \multicolumn{3}{c}{Metrics} \\ 
& CR  $\uparrow$ & IV [$cm^3$] $\downarrow$ & ID $[mm] \downarrow$ \\ \hline
D-Grasp~\cite{D-Grasp} & 0.055 & \textbf{5.56} & \textbf{8.60}   \\
Ours w/o SDF  & 0.073 & 10.79 & 11.42 \\ 
Ours w/ object-relative pose & \underline{0.077} & 9.38 & 11.41   \\
Ours w/ first frame relative pos. (Ours) & \textbf{0.086} & \underline{9.03} & \underline{11.39}   \\
\hline
\end{tabular}
}

\label{tab:pose_repr_ablation}
\end{table}

\begin{table}[h!]
  \centering
      \caption{\textbf{Text evaluation.} 
    We demonstrate that detailed text descriptions enable us to generate motions more representative of the description, and allow fine-grained controllable hand-object motion synthesis. 
    } 
    \resizebox{0.9\linewidth}{!}{
        \begin{tabular}{@{}cccccccc@{}}
        \toprule
        \multicolumn{1}{c}{Train} & \multicolumn{1}{c}{Test}             &\multicolumn{1}{c}{AR} &\multicolumn{4}{c}{Hand correctness} & Cosine   \\
         Input    & Input    &  & Right & Left & Both & Total & Similarity \\ 
        \midrule
        \begin{tabular}[c]{@{}c@{}}simple \end{tabular} & \begin{tabular}[c]{@{}c@{}}simple\end{tabular}  &      0.837 &  n/a     &  n/a   &  n/a    & n/a & 0.72   \\
        \begin{tabular}[c]{@{}c@{}}detailed \end{tabular} & \begin{tabular}[c]{@{}c@{}}simple\end{tabular}    &    0.516    &  n/a     &  n/a   &  n/a    & n/a & 0.43  \\
        \begin{tabular}[c]{@{}c@{}}simple \end{tabular} & \begin{tabular}[c]{@{}c@{}}detailed \end{tabular} &    0.702   &  0.709     & 0.111    & 0.0  & 0.593 & 0.43 \\ 
        \begin{tabular}[c]{@{}c@{}}detailed \end{tabular} & \begin{tabular}[c]{@{}c@{}}detailed \end{tabular} & \textbf{ 0.887} & \textbf{0.869}  & \textbf{0.862} & \textbf{0.75} & \textbf{0.865} & 0.66  \\
        \bottomrule
        \end{tabular}
    }

  \label{tab:textual_augmentations}
\end{table}


\section{Discussion and Conclusion}
We have introduced a framework to generate plausible hand-object interactions from textual descriptions. Specifically, we have proposed a two-stage diffusion framework that decouples sequences into grasping and interaction, and can use single grasps as guidance to the diffusion model to improve generalization to unseen objects. While we provide a first step towards HOI synthesis generalization, there are limitations of our framework. To remove physical artifacts, physics could be explicitly integrated into the diffusion process \cite{yuan2023physdiff}. The inference time is still relatively slow. To increase efficiency, operating in latent space is an interesting direction to explore \cite{chen2023mld}. 


\bibliographystyle{ACM-Reference-Format}
\bibliography{main}

\appendix
\clearpage


\section{Implementation Details}
\label{supp:impl}

While we experiment with a transformer backbone as proposed in MDM \cite{tevet2023mdm}, \methodname's~main architecture is based on UNet with Adaptive Group Normalization (AdaGN) which was originally proposed in~\cite{dhariwal2021diffusion} and also adapted in~\cite{karunratanakul2023gmd} for sequence prediction tasks. We adapt this network architecture for our two-stage design, that involves grasping and interaction diffusion models. We provide the hyperparameter settings used in our architecture in Table~\ref{tab:related_work}.

\section{Experimental Details}
\label{supp:exp_details}



\subsection{Obtaining Grasp References} 
\label{sec:app_references}
As mentioned in the main part of the paper, we show three different ways grasp references can be obtained, namely via existing datasets, pose estimation, or grasp synthesis. Note that these procedures for acquiring hand pose references have already been established in related works \cite{D-Grasp, xu2023unidexgrasp}, albeit for the sole purpose of grasping and not interaction.

\paragraph{Hand-Object Datasets} The most straightforward way to attain grasp references is to make use of existing hand-object datasets (see Sec. 2). There are some single frame datasets like ObMan \cite{hasson19_obman} or AffordPose \cite{ye2023affordance} that could be directly used as grasp references. However, most existing datasets contain sequential data, hence, we need to identify which frames are potential candidates for hand-object grasps. In our paper, we run guidance experiments on our test split \updatered{of} the GRAB dataset \cite{GRAB}. To acquire suitable frames for guidance, we apply the heuristics described in Sec. 4.2.1. In this case, we can directly use the preprocessed data from GRAB as single keyframes for guidance.

\paragraph{Image-Based Pose Estimation}
For this experiment, we need to extract the grasp references from images. We run an image-based pose estimator~\cite{pavlakos2024reconstructing} on single images from HO3D~\cite{hampali2020ho3d} to obtain object-relative hand poses. In this experiment, we focus on extracting a single hand pose from the image. Please note that we assume the object mesh to be known and at a normalized position to be suitable for our experiments. Given the object-relative hand pose, we need to make three modifications to convert it to the same input space our model has. First, we have to compute the features for our approximated signed-distance space by finding the closest vertex on the object mesh for each joint position. Second, we convert the hand pose from joint angle space to MANO's PCA space. Lastly, because the GRAB dataset is normalized to approach objects from one direction, we have to adjust the orientation of the \updatered{hand and object} to roughly be in line with this direction. This could be alleviated by adding data augmentation to the training data in the future. We also leave exploring object mesh reconstruction for future work, for example, by utilizing methods that reconstruct the object mesh along with the hand pose estimation \cite{fan2024hold}.

\paragraph{Grasp Synthesis}
In this experiment, we use a grasp synthesis method, GraspXL \cite{zhang2024graspxl}, to obtain grasp references for guidance. To this end, we run GraspXL on several objects to obtain grasping motions. As we only require one frame for the guidance, we take the last frame of the grasping motion as the keyframe. Since GraspXL allows conditioning on approaching direction, we select grasps whose approaching direction aligns with the GRAB dataset. Again, we compute the approximated signed distance by finding the closest vertex on the object mesh for each joint position and convert the hand pose from joint angle to MANO PCA space. There is no further conversion needed, as the object poses are already normalized. Similar to the experiment with hand poses from images, we only acquire an object-relative grasp and its respective features for one hand.

\subsection{Additional Motion Feature Evaluation Metrics.}
\label{supp:mf_metrics}
For completion, we also provide metrics based on an action recognition classifier following previous motion synthesis works \cite{IMoS, tevet2023mdm}.

\paragraph{\textbf{Action Recognition Accuracy (AR).}} Given a pretrained classifier, we input motions generated by \methodname~as well as our baselines and report top-1 percentage accuracy. The recognition accuracy indicates the correlation of the action type and the motion.

\paragraph{\textbf{Diversity (DIV).}} We compute the diversity score by computing the variance of the features extracted from the action classifier across all action categories. Given a set of features extracted from generated motions across all action categories, two subsets with the same size, N, are randomly sampled. These features are denoted as $\{ \bf{v}_1,\bf{v}_2, \dots, \bf{v}_N \}$ for the first subset and $\{ {\bf{v}_1}^{'},{\bf{v}_2}^{'}, \dots, {\bf{v}_N}^{'} \}$ for the second subset. With $N=200$, the diversity is defined as

\begin{equation}
\textrm{DIV} = \frac{1}{N} \sum_{i=1}^{N} || \bf{v_i} - \bf{{v_{i}^{'}}} || 
\end{equation}

\paragraph{\textbf{Multimodality (MM).}} Different from diversity, multimodality computes variance only within a specific action. An overall score is attained based on averaging the variances across all action types.

\paragraph{\textbf{Fr\'echet Inception Distance (FID).}}  FID is computed based on the (Fr\'echet) distance of the features of real ground-truth motions and generated motions. It has the assumption that the features computed from the real data and synthesized data have a Gaussian distribution. Earlier work demonstrated that for small sample sizes, FID metric is biased and not stable \cite{noguchi2019image}. 

\paragraph{\textbf{Kernel Inception Distance (KID).}} KID metric relaxes the Gaussianity assumption and aims to improve upon FID. It measures the squared Maximum Mean Discrepancy (MMD) between the feature representations of the real and generated samples using a polynomial kernel. This metric has been found to be more robust when the sample size is small~\cite{noguchi2019image}.

\subsection{Metrics - Training the Action Classifier.} 
\label{supp:classifier}
To provide motion feature metrics (see Sec.~\ref{supp:mf_metrics}) based on an action classifier, we train a standard RNN action recognition classifier on the GRAB
dataset. We use the final layer of the classifier as the motion feature
extractor for calculating action recognition accuracies as well as diversity (DIV) and multimodality (MM) scores, following~\cite{IMoS} and~\cite{guo2020action2motion}. We employ the online available code from IMoS and use the same network architecture and parameters to train our model. 

\begin{figure}
  \includegraphics[width=1.0\linewidth]{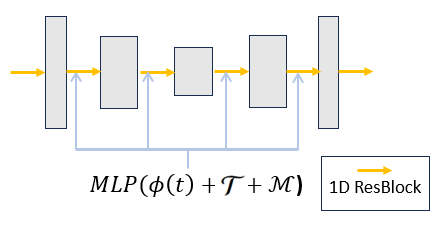}
  \caption{\textbf{Overview of the diffusion architecture.} Our pipeline relies on a UNet block and  processes three input signals:
the time step $\phi(t)$, a text-prompt embedding $\mathcal{T}$ and an object shape encoding $\mathcal{M}$. The time step is encoded using sinusoidal functions, the text-prompt embedding is generated by the CLIP text encoder model and the object encoding is obtained from BPS\cite{prokudin2019efficient}. Similarly to~\cite{karunratanakul2023gmd}, we use Adaptive Group normalization in 1D block}
  \label{fig:unet_block}
\end{figure}
\begin{table}
\centering
\begin{center}
\resizebox{1.0\linewidth}{!}{%
\begin{tabular}{l|cc}
   \toprule
                         & Grasping  & Interaction \\
    Parameter            & Model      & Model  \\
    \midrule
    Batch size           & 32         & 32            \\
    Base channels        & 256        & 512            \\
    Latent dimension   & 256        & 512          \\
    Channel multipliers  & (1, 1, 1)  & (2, 2, 2, 2) \\
    $\beta$ scheduler      &     \multicolumn{2}{c}{Cosine \cite{nichol2021improved}}           \\
    Learning rate        &     1e-4       &     1e-4        \\
    Optimizer            &         \multicolumn{2}{c}{AdamW (wd = 1e-2)}            \\
    Training $T$         &   \multicolumn{2}{c}{400}      \\
    Diffusion loss       & $\epsilon$ prediction  & $\bf{x}_{0}$ prediction \\
    Diffusion var.       &     \multicolumn{2}{c}{Fixed small $\tilde{\beta}_t=\frac{1-\alpha_{t-1}}{1-\alpha_t}\beta_t$}           \\
    Model avg. beta      &      \multicolumn{2}{c}{0.9999}     \\
   \bottomrule
\end{tabular}
}
\end{center}
\caption{\heading{Network architecture} Model and  hyperparameters of \methodname}
\label{tab:related_work}
\end{table}

\renewcommand{\thefootnote}{\textasteriskcentered}

\update{
\subsection{Annotation Examples}
\updatered{We provide a few examples of our textual annotations in Table \ref{tab:annotations}.} Please see our code release for the entire set of annotations\updatered{\footnote{\updatered{The code and data will be available here: \url{https://diffh2o.github.io}}}}.}

\subsection{Visualization Pipeline.} We use AITViewer~\cite{Kaufmann_Vechev_aitviewer_2022} to visualize motions synthesized by \methodname~as well as our baselines. IMoS synthesizes 15 frames of human-object interactions. They  determine the hand which is in contact with the object according to GT, which can lead to a jump of the object from one hand to the other for certain actions. To make the frame rates comparable with \methodname, we upsample the synthesis results of IMoS to 75 frames using spherical interpolation. Therefore, this jump may result in an artifact in which the object is floating from one hand to another.

\section{Additional Experiments}

\subsection{Action Recognition Experiments}

\paragraph{\textbf{Training action classifier on the whole GRAB data.}} IMoS \cite{IMoS}  reports results using the whole dataset to train the action classifier on ground-truth motions and testing on the synthesized actions. Since their model initializes the sequences with ground-truth from the test data, this could positively bias their scores. Therefore, we omit test data from training, in our analysis in Table 1 of the main paper, both for our approach and IMoS. We show however in this supplemental material (Table~\ref{tab:all_training_data}) that our model is still able to outperform IMoS even when all the GRAB data, including test data, is included in training the classifier. This is a setup which is more favorable for IMoS as the first frames of synthesized IMoS motion is taken from the test data. This re-confirms the ability of our method to encode diverse motion features that represents realistic hand-object interactions. 
Furthermore, we train two models, one that uses only hand 3D joints as input and another incorporating object 3D position as additional input. We demonstrate that \methodname's synthesized object motions outperform IMoS in both settings. This confirms the benefits of jointly modeling hand-object motion instead of decoupling it into separate components. Note also that as expected, the classifier can correctly identify all test sequences of the real mocap data, since it has seen the test data during training in the IMoS setting.

\paragraph{\textbf{FID and KID metrics.}} It's been shown by earlier studies~\cite{noguchi2019image},~\cite{jayasumana2023rethinking},~\cite{chong2020effectively} that  Fr\'echet Inception Distance (FID) metric is biased and not stable for calculating the distances for small datasets. Since the GRAB  subject split only contains 144 motions, in addition to FID, we also report the more stable Kernel Inception Distance (KID) metric  which provides more reliable scores with fewer samples (Table~\ref{tab:fid_stds}). Table~\ref{tab:fid_stds} also includes the standard deviations  for the multimodality and diversity metrics based upon 20 repetitions of the evaluation.

\begin{table*}[t!]
  \centering
    \resizebox{0.85\textwidth}{!}{%
  \begin{tabular}{@{}l|c|ccccccc@{}}
    \toprule
      &Method           & \begin{tabular}[c]{@{}c@{}}Hand AR ($\uparrow$)\end{tabular}   & \begin{tabular}[c]{@{}c@{}} Hand+Object AR ($\uparrow$)\end{tabular}  &  DIV ($\rightarrow$)      & MM ($\rightarrow$)  & KID ($\downarrow$) & FID ($\downarrow$) \\ \midrule\midrule
    \parbox[t]{11mm}{\multirow{3}{*}{\rotatebox[origin=c]{90}{ \begin{tabular}[c]{@{}c@{}} \hspace{-1.8mm} Unseen\\ \hspace{-2mm} subject \\ \hspace{-2mm} split \end{tabular}}}} 
    & Real Mocap       & 1.0                                                                             & 1.0                                                                                      & 1.1358$\pm$0.0164         &  0.3105$\pm$0.0163            & -                  &  -    \\
    & IMoS~\cite{IMoS}   & 0.7017                                                                          & 0.6754                                                                                   & 1.1053$\pm$0.0128         &  0.2882$\pm$0.0114            & 0.005811           & \textbf{0.6267} \\ 
    & DiffH$_{2}$O       & \textbf{0.8125}                                                                 & \textbf{0.9028}                                                                          & \textbf{1.1440$\pm$0.0115} &  \textbf{0.3176$\pm$0.0127} & \textbf{0.005346}   & 0.8342  \\ 
    \midrule
    \parbox[t]{11mm}{\multirow{3}{*}{\rotatebox[origin=c]{90}{\begin{tabular}[c]{@{}c@{}} \hspace{-1.3mm}Unseen\\ \hspace{-2mm} object \\ \hspace{-2mm} split\end{tabular}}}} 
    & Real Mocap       & 1.0                                                                             & 1.0                                                                                      & 1.1324$\pm$0.0119         & 0.2355$\pm$0.0044             & -                  & -       \\ 
    & IMoS*~\cite{IMoS}  & 0.7161                                                                          & 0.8000                                                                                   & 1.1173$\pm$0.0112         & \textbf{0.2513$\pm$0.0104}    & 0.006038 & \textbf{0.7101}    \\ 
    & DiffH$_{2}$O      & \textbf{0.7836}                                                                  & \textbf{0.8125}                                                                          &\textbf{1.1225$\pm$0.0120} & 0.2174$\pm$0.0074             & \textbf{0.005494} & 0.8821    \\  
    \bottomrule
  \end{tabular}
  }
  \caption{\textbf{Action feature based metrics using all training data.} We report action feature based metrics using action recognition models trained on all of the GRAB training data following the protocol of~\cite{IMoS}. We either use a subject-based split (top 3 rows) or an object-based split (bottom 3 rows). $\downarrow$ denotes lower values are better, and $\rightarrow$ denotes values closer to the ground-truth are  better. Our results achieves state-of-the-art accuracy across different metrics.}
  \label{tab:all_training_data}
\end{table*}

\begin{table*}[t!]
  \centering
    \resizebox{0.65\textwidth}{!}{%
  \begin{tabular}{@{}l|c|ccccc@{}}
    \toprule
      &Method            &  DIV ($\rightarrow$)   & MM ($\rightarrow$)  & KID ($\downarrow$) & FID ($\downarrow$) \\ \midrule\midrule
    \parbox[t]{11mm}{\multirow{3}{*}{\rotatebox[origin=c]{90}{ \begin{tabular}[c]{@{}c@{}} \hspace{-1.8mm} Unseen\\ \hspace{-2mm} subject \\ \hspace{-2mm} split \end{tabular}}}} 
    & Real Mocap       & 1.0632$\pm$0.0131      &  0.2155$\pm$0.0091  & -                  &  -    \\
    & IMoS~\cite{IMoS}   & 1.0237$\pm$0.0158      &  \textbf{0.2538$\pm$0.0103}  & 0.008065           & 0.6267 \\ 
    & DiffH$_{2}$O       & \textbf{1.0757$\pm$0.0129}      & 0.3037$\pm$0.0079  & \textbf{0.006697}                   & 0.8342  \\ 
    \midrule
    \parbox[t]{11mm}{\multirow{3}{*}{\rotatebox[origin=c]{90}{\begin{tabular}[c]{@{}c@{}} \hspace{-1.3mm}Unseen\\ \hspace{-2mm} object \\ \hspace{-2mm} split\end{tabular}}}} 
    & Real Mocap      & 1.0757$\pm$0.0129      & 0.2002$\pm$0.0063   & -                  & -       \\ 
    & IMoS*~\cite{IMoS}  & 1.0471$\pm$0.0129      & \textbf{0.2202$\pm$0.0093}   & 0.008641           & 0.6593   \\ 
    & DiffH$_{2}$O       & \textbf{1.0942$\pm$0.0125}      & 0.2307$\pm$0.0060   & \textbf{0.007503}           & 0.8425   \\ 
    \bottomrule
  \end{tabular}
  }
  \caption{\textbf{Details of the quantitative analysis with action feature based metrics.} We provide further details for the quantitative analysis in Table 1 of the main paper and report standard deviations of multimodality and diversity metrics as well as FID and KID scores. The results here are obtained using action recognition models trained on hand pose data of the respective training splits as indicated in the first column. We either use a subject-based split (top 3 rows) or an object-based split (bottom 3 rows). For IMoS, we use the same pretrained model, which is trained on unseen subject split, across all our experiments (unseen subject and unseen object splits) due to difficulties in reproducing training performance for the unseen object split (indicated with a * in the table, IMoS*). Therfore IMoS* sees a part of the unseen object test split during training the model which positively biases their score. $\downarrow$ denotes lower values are better, and $\rightarrow$ denotes values closer to the ground-truth (real mocap) are  better. Our results achieves state-of-the-art accuracy across different metrics. }
  \label{tab:fid_stds}
\end{table*}

\paragraph{\textbf{Full-sequence training.}} Given an initial pose of the grasp moment, IMoS only synthesizes interaction sequences. To compare our results to IMoS in Table 1 of the main paper, we evaluated our models only on the interaction sequence part of the synthesis. In Table~\ref{tab:full_sequence}, we also provide diversity, multimodality, FID, KID and action recognition scores computed over the full sequence including grasp and interaction stages. We compare our results against motions from motion capture and observe that we can achieve scores similar to real mocap, which demonstrates our capability to synthesize realistic hand-object interactions.

\begin{table*}[t!]
  \centering
    \resizebox{0.85\textwidth}{!}{%
  \begin{tabular}{@{}l|cccccccc@{}}
    \toprule
      Method           & \begin{tabular}[c]{@{}c@{}}Hand Act. \\ Rec. Accuracy ($\uparrow$)\end{tabular}   & \begin{tabular}[c]{@{}c@{}} Hand+Object Act.\\ Rec. Accuracy ($\uparrow$)\end{tabular}  &  DIV ($\rightarrow$)      & MM ($\rightarrow$)  & KID ($\downarrow$) & FID ($\downarrow$) \\ \midrule\midrule

    Real Mocap (Full-sequence)                      & 1.0                                                              & 1.0                                                                                      & 1.0983$\pm$0.0059         & 0.2235$\pm$0.0042             &-       &-    \\         
    DiffH$_{2}$O (Full-sequence)      & 0.7720                                                           & 0.8031                                                                         &1.0948$\pm$0.0120 & 0.1555$\pm$0.0057            & 0.0067 & 0.7993   \\ 

    \bottomrule
  \end{tabular}
  }
  \caption{\textbf{Action feature based metrics on full sequence synthesis.} We report action feature based metrics on our unseen object split using action recognition models trained on full sequences instead of only post-grasp data as in~\cite{IMoS}. $\downarrow$ denotes lower values are better, and $\rightarrow$ denotes values closer to the ground-truth are  better. Our results achieves state-of-the-art accuracy across different metrics.}
  \label{tab:full_sequence}
\end{table*}

\begin{table*}[h!]
    \centering
    \resizebox{0.75\textwidth}{!}{%
    \update{
    \begin{tabular}{|p{3cm}|p{12cm}|}
        \hline
        \textbf{Object Name} & \textbf{Annotation} \\
        \hline
        camera & The person uses their left hand to pick up the camera, brings it closer to their eyes to take a photo, and then places it back on the table with their right hand. \\
        \hline
        stapler & The person repeatedly picks up and places the stapler on the table always with their right hand. \\
        \hline
        mouse & The person picks up the mouse with their right hand, passes it to someone on their right below chest level, and then places it back on the table with their right hand. \\
        \hline
        alarm clock & The person repeatedly picks up the alarm clock with their right hand, lifts it, and places it back on the table with their right hand. \\
        \hline
        duck & Using the right hand, the person picks up the duck, examines it closely, and then places the duck back down on the table. \\
        \hline
        bowl & The person picks up the bowl with both hands, drinks from it, and places it back on the table with both hands. \\
        \hline
        headphones & The person picks up the headphones with their left hand, puts them on with both hands, and places them on the table with both hands. \\
        \hline
        banana & The person picks up the banana with their right hand, passes it to their left hand, then hands it to someone on their left side at shoulder level with their left hand, and finally places it back on the table with their right hand. \\
        \hline
    \end{tabular}
    }
    }
    \caption{Examples of our textual sequence descriptions}
    \label{tab:annotations}
\end{table*}

\subsection{Analysis on Generalizability}
\label{sec:analysis_gen}
When we analyze per-unseen-object performance on GRAB, our method effectively generalizes to unseen objects with various sizes and shapes (e.g., apple, clock, train, elephant, mug, etc.). We also evaluated the influence of object scale on three test objects by using objects of the same shape but different scales during training: our model performs well on medium cylinder and large torus (584cm³), but faces challenges on small pyramid (64cm³, video: 04.40, low contact-ratio). The average bounding-box volumes for our train and test sets are 741cm³ and 625cm³, respectively. In conclusion, our model adapts well to unseen shapes and a wide range of sizes, but largely out-of-distribution sizes (e.g., also \text{large\_cracker\_box} from HO3D: 2208cm³) lead to less favorable physics metrics. 

\subsection{Discussion on Grasp References}
Here we provide a more detailed discussion about the role of grasp references that are used for guidance. 

\paragraph{\updatered{Strengths:}} Grasp guidance is a technique to increase the controllability and improve generalization on unseen objects if grasp references are available. Note, however, that our model without guidance already significantly outperforms baselines (cf. Table 1 in the main paper), making it an ideal tool for potential downstream use cases. \updatered{Grasp} guidance yields better generalization (e.g., less interpenetration as shown in Table 2 in the main paper), because it provides a signal to the diffusion model about how an object should be grasped. Especially for unseen object geometry, this can serve as important information about feasible grasping poses of the hand. One of the other main advantages of grasp guidance is that it allows defining the grasping hand(s) on the object, which may be important if rich textual annotations such as the ones presented in Sec. 4.4 are not available. 

\paragraph{Weaknesses and Failures:} Grasp guidance faces an important trade-off: while the signal can be helpful for object generalization and controllability of handedness, it requires access to single hand-object poses. We detail in \secref{sec:app_references} how such grasp references can be obtained. Hence, there are conversion steps that are necessary to make guidance work on unseen object geometry, which require some manual effort. For instance, due to the small dataset size of GRAB and the similar approaching direction for all objects, an alignment step is necessary to make the guidance signal useful. In the future, larger dataset distributions or additional data augmentation techniques may help alleviate this. Furthermore, as our model is conditioned on both the text description and the grasp guidance, one needs to make sure that the two align: for example, if the grasp reference shows grasping a cup from above, where the prompted action is "drink from the cup", the model is forced to select which signal to follow. In practice, as the guidance is only a weak guidance signal at a single frame, the model will favor the text description over the guidance. Another limiting factor is the size of the unseen objects as discussed in \secref{sec:analysis_gen}. For example, if the object is too large, the target grasp may be reached in the grasping phase, but after imputing the grasping into the interaction stage, the model may neglect the out-of-distribution grasp and output a grasp that displays physical artifacts such as interpenetration. \\

Overall, grasp guidance is a technique that shows the potential to provide a link between how an object should be grasped and subsequent actions that can be prompted via text. While there are limiting factors in its current state, i.e., manual alignment and matching text prompt with the grasp reference, future work could focus on enhancing existing datasets and using improved guidance techniques \cite{karunratanakul2024dno} to further increase performance.

\section{Computational Resources} 
The training of our model takes approximately 48 GPU hours on a single NVIDIA V100. Our model has a throughput of 32 samples per second. The total inference time for 32 samples is approximately 300 seconds.

\section{Ethical Discussion} 
Our method uses a generative model, namely diffusion models, to synthesize realistic hand-object motions. While the generated outputs at this point are in 3D and not rendered on photorealistic images with natural backgrounds, such a feature could be added in the future. This may lead to malicious use of our research, such as generating virtual \updatered{deepfakes}. Other downstream applications that could have problematic \updatered{use cases} are robotics, where our generated data could be used to train robots in handling objects in a desired way. For example, robots could be taught how to handle weapons. \updatered{Moving} forward with the development of our method, we hope that openly discussing the technical details, implementation, and sharing our code will ensure that the technology is well comprehended, accessible to the public and measures to prevent and flag malicious uses are easier to implement.

\end{document}